\pdfoutput=1

\documentclass[11pt]{article}

\usepackage[]{emnlp2021}

\usepackage{times}
\usepackage{latexsym}

\usepackage[T1]{fontenc}

\usepackage[utf8]{inputenc}

\usepackage{microtype}

\usepackage{booktabs}
\usepackage{graphicx}
\usepackage{subcaption}
\usepackage{xspace}

\usepackage{amsfonts}
\usepackage{amsmath}
\usepackage{enumitem}

\usepackage{inconsolata}
\usepackage{balance}

\usepackage{todonotes}
\makeatletter
\newcommand*\iftodonotes{\if@todonotes@disabled\expandafter\@secondoftwo\else\expandafter\@firstoftwo\fi}  
\makeatother

\usepackage{euflag}

\usepackage{cleveref}
\crefname{section}{\S}{\S\S}
\crefname{table}{Table}{}
\crefname{figure}{Figure}{}
\crefname{algorithm}{Algorithm}{}
\crefname{equation}{Eq.}{}
\crefname{appendix}{App.}{}
\crefname{prop}{Proposition}{}
\crefname{thm}{Theorem}{}
\crefformat{section}{\S#2#1#3}  

\newcommand{\vl}{\textsc{V\&L}\xspace}

\newcommand{\faster}{Faster R-CNN}
\newcommand{\entities}{Flickr30k Entities}
\newcommand{\entsvg}{LabelMatch}

\newcommand{\object}{object\xspace}
\newcommand{\phrase}{phrase\xspace}
\newcommand{\image}{image\xspace}
\newcommand{\sentence}{sentence\xspace}

\newcommand{\objabl}{\textbf{Object} ablation\xspace}
\newcommand{\phrabl}{\textbf{Phrase} ablation\xspace}
\newcommand{\allabl}{\textbf{All} ablation\xspace}
\newcommand{\noneabl}{\textbf{None} ablation\xspace}

\title{Vision-\emph{and}-Language or Vision-\emph{for}-Language?\\On Cross-Modal Influence in Multimodal Transformers}

\usepackage{tipa}
\newcommand{\unitn}{\text{\normalfont \textipa{D}}}
\newcommand{\ku}{\text{\normalfont  \textipa{C}}}
  
\usepackage{lipsum}
\newcommand\blfootnote[1]{%
  \begingroup
  \renewcommand\thefootnote{}\footnote{#1}%
  \addtocounter{footnote}{-1}%
  \endgroup
}

\author{
  Stella Frank$^{\ast,\unitn}$ \ \ 
  Emanuele Bugliarello$^{\ast,\ku}$ \ \  
  Desmond Elliott$^{\ku}$ \\
  $^{\unitn}$University of Trento \ \ 
  $^{\ku}$University of Copenhagen \\
  \href{mailto:stella.frank@unitn.it}{stella.frank@unitn.it} \ \
  \href{mailto:emanuele@di.ku.dk}{\{emanuele}, \href{mailto:de@di.e.dk}{de\}@di.ku.dk}
}

\begin{document}

\maketitle

\begin{abstract}
Pretrained vision-and-language BERTs aim to learn representations that combine information from both modalities. We propose a diagnostic method based on \emph{cross-modal input ablation} to assess the extent to which these models actually integrate cross-modal information. This method involves ablating inputs from one modality, either entirely or selectively based on cross-modal grounding alignments, and evaluating the model prediction  performance on the other modality. Model performance is measured by modality-specific tasks that mirror the model pretraining objectives (e.g.~masked language modelling for text). Models that have learned to construct cross-modal representations using both modalities are expected to perform worse when inputs are missing from a modality. We find that recently proposed models have much greater relative difficulty predicting text when visual information is ablated, compared to predicting visual object categories when text is ablated, indicating that these models are not symmetrically cross-modal.\\
\blfootnote{$^*$Equal contribution.} 
\end{abstract}

\section{Introduction}

\begin{figure}[t]
    \includegraphics[width=\linewidth, trim={3.85cm 12.7cm 4cm 6.9cm}, clip]{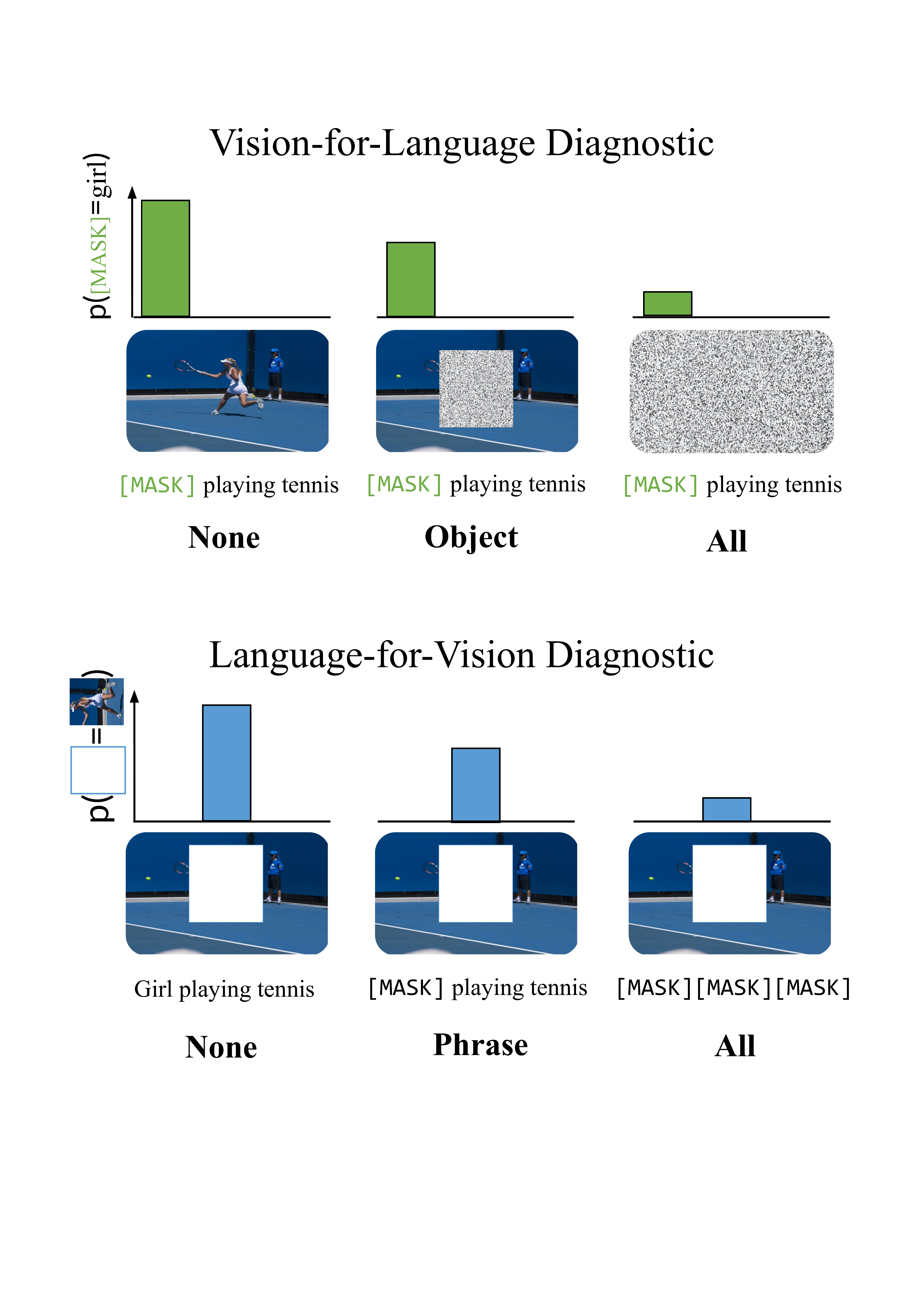}
    \label{fig:sketch}
    \vspace{-.5cm}
    \caption{Cross-modal input ablation tests how well models predict masked data, given ablated inputs in the other modality. The Vision-for-Language Diagnostic (top) measures the effect of ablation, of either the aligned object or the full image, on masked token prediction, while the Language-for-Vision Diagnostic (bottom) measures the effect of ablating either the aligned Phrase or the entire sentence when predicting the properties of a masked image region.} 
    \label{fig:intro}
\end{figure}

Vision-and-language (V\&L) BERT models extend the BERT architecture
\cite{devlin-etal-2019-bert} to produce cross-modal contextualised
representations of multimodal inputs \cite{NEURIPS2019_c74d97b0,tan2019lxmert,chen2020uniter,Su2020VL-BERT:}.  These models have proven to be highly effective when fine-tuned for a range of downstream tasks. 
However, in spite of their versatility, little is known about how these models use cross-modal information:
do their learned representations for language tasks include visual information (vision-for-language) and vice-versa (language-for-vision)?
\newcite{Su2020VL-BERT:} claim that their model will use visual context for predicting masked words, thus aligning visual and linguistic contexts, while \newcite{Lu_2020_CVPR} hope to force their model to rely more heavily on language to predict image content.
The extent to which these statements hold is unknown,
in part because of the difficulty of analysing exactly how these models use information across modalities.

In this paper, we introduce a \emph{cross-modal input ablation} method to quantify the degree to which a pretrained model has learned to use cross-modal information.
This method, which requires no additional training, involves ablating all or some of the inputs from one modality when making a prediction in the other modality, and measuring the change in performance.
See Figure~\ref{fig:intro} for an illustrative sketch.
Performance is measured using the same masked-target prediction tasks used during pretraining,
in which a model must predict either a masked token given surrounding text and visual context (masked language modelling), or a masked visual object given the surrounding visual context and accompanying text (masked region classification).
Cross-modal input ablation thus captures the degree to which a model depends on cross-modal inputs and activations when generating predictions.

Our use of input ablation to assess cross-modal recruitment is novel.
Previous analyses of multimodal models have used diagnostic classifiers and attention analyses
\cite{li-etal-2020-bert-vision,cao-etal-2020-behind,parcalabescu-etal-2021-seeing}.
In comparison, cross-modal input ablation has the following advantages:
\begin{itemize}[noitemsep, topsep=1pt]
  \item It is straightforward to perform, and easy to interpret, requiring no intervention in the model and only minimal intervention on the data.
  \item As an intrinsic diagnostic, it examines the model directly, unlike methods that add learned parameters, such as classifier-based probing approaches \cite{hupkes2018visualisation}.
  \item It does not require interpreting activations or attention, which can be difficult \cite{jain-wallace-2019-attention}.
\end{itemize}
We envision cross-modal input ablation as a useful diagnostic check to be performed during model development, to test the effect of changes in architecture and optimisation.

In this paper we perform a case study of cross-modal input ablation on existing models, to demonstrate its utility in understanding model behaviour.
We test models that have different architectures but the same initialisation and training procedures~\cite{bugliarello-etal-2021-multimodal}.
Our cross-modal input ablation results show that these models do learn to use cross-modal information, resulting in multimodal representations, but this is not equally true across both modalities.
In particular, the representations of text segments are strongly influenced by the visual input, while the representations of visual regions are much less influenced by the accompanying textual input.
This indicates that the level of cross-modal information exchange is \emph{not symmetrical}: the models have learned to use vision-for-language more than language-for-vision.

In subsequent analyses, we attempt to understand the lack of recruitment of language-for-vision, in order to identify possible avenues for improvement.
Our experiments investigate different loss functions, initialisation and pretraining strategies, and visual co-masking procedures. None of these factors changes model behaviour significantly.
However, we find that the visual object annotations used in pretraining, which are automatically generated by an object detector, are much noisier than expected.
We surmise that the ensuing models do not recruit text because it is not useful for predicting these noisy object features, and discuss implications for future \vl models.

\section{Related Work}
Significant efforts have been devoted to understanding what is learned by text-only pretrained language models~\cite{rogers-etal-2020-primer}, which can be categorised as probing based on diagnostic classifier, analysing attention weights, or direct evaluations~\cite{belinkov-glass-2019-analysis}.
However, much less work has looked at \vl pretraining.
Two studies have examined learned attention weights:
\newcite{li-etal-2020-bert-vision} find that there are attention heads specialising in entity phrase grounding in VisualBERT, but do not investigate text-to-vision attention.
\newcite{cao-etal-2020-behind} find that the textual modality dominates in UNITER and LXMERT, both in terms of overall attention and more specifically during visual co-reference resolution tasks.
In a direct evaluation analysis,
\newcite{parcalabescu-etal-2021-seeing} evaluate VilBERT and LXMERT on a counting task and find more evidence of dataset bias (predicting frequent numbers) than accurate image grounding. 
These results all indicate that cross-modal interaction within these models may not be as strong or as symmetric as often assumed, which is further confirmed by the methods introduced in this paper. In addition, we evaluate a larger set of models, covering both dual-stream and single-stream models.

The ablation method introduced here is related to concurrent work in language modelling~\cite{oconnor-andreas-2021-context} and machine translation~\cite{fernandes2021measuring}, in which the effect of removing context is measured. \newcite{oconnor-andreas-2021-context} and~\newcite{fernandes2021measuring} use ablations at both training and evaluation, whereas we only study the effect of ablating inputs during evaluation.

\section{Proposed Approach}\label{sec:approach}

We use ablations to determine whether pretrained vision-and-language models combine
inputs from both modalities when making predictions.
In general, ablations are used to test the hypothesis that a certain model component (here, multimodal inputs) is responsible for certain model behaviour:
ablated models are expected to perform differently because of the missing component.
Importantly, this hypothesis is falsifiable~\cite{popper34logik} if the ablation leads to no change.

Multimodal models are hypothesised to use cross-modal activations, triggered by multimodal inputs, when making predictions.
If a multimodal model relies on activations from certain input data, e.g.~object category labels, to make predictions, the ablation of
this input will lead to a change in performance, whereas if the model has not learned to use the input, removing it will have no effect.
The predictions we evaluate are the same kinds of predictions that the models have learned to make during pretraining,
i.e.~predicting the identity of masked tokens or regions given (possibly ablated) multimodal text and image contexts.

Input data are in the form of an \emph{\image} paired with a \emph{\sentence} describing the image; within the \sentence, \emph{phrases} can refer to particular \emph{objects} in the image.
We expect such aligned, grounded, phrase--object pairs to elicit especially strong cross-modal activations at prediction time.
By ablating the alignment link, we test models' ability to create and use grounded alignments.
We also test whether non-specific cross-modal information is used, by ablating whole modalities.

\subsection{Vision-for-Language Diagnostic}

The language task consists of predicting masked tokens, possibly using visual inputs.
For visual input ablation, we compare the following setups:

\begin{description}[noitemsep,topsep=1pt]
	\item[None:] None of the visual features are ablated, i.e.~the model has access to the full image. This is the original multimodal setting and thus where a model that uses multimodal information effectively should perform best.
	\item[Object:] Here we remove only the image regions that correspond to the aligned textual phrase. This tests a model's ability to ground text to specific regions of the image, by breaking any possible alignment. However, the model can still use surrounding visual context features.
	\item[All:] All the visual features are ablated and the model needs to predict masked textual tokens from its textual context only. Models that depend on multimodal inputs should suffer.
\end{description}

\cref{fig:intro} (top, middle) shows an example of \objabl{}, where the target is to predict ``\textit{girl}'' and the ablated visual input is the image region corresponding to the tennis player.

\subsection{Language-for-Vision Diagnostic}
The vision task is to predict the target object category within a specific region of the image, possibly aided by the text caption.
In this case, a single region, corresponding to an \object aligned to a \phrase in the sentence, is selected as the target.
The textual input is ablated analogously to the visual input:

\begin{description}[noitemsep,topsep=1pt]
	\item[None:] None of the text is ablated, i.e.~the model sees the entire sentence.
  \item[Phrase:] Here we only ablate the tokens in the \phrase aligned to the target \object.
	\item[All:] All tokens are ablated and the model is required to predict the masked visual region from its visual context only.
\end{description}

\cref{fig:intro} (bottom, middle) shows an example of \phrabl{}, which tests the extent to which the model relies on grounding information when predicting the target tokens.
If ablating the \phrase does not change performance compared to ablating the entire sentence, then the model has not learned to use \phrase-to-\object alignments to infer what is missing in the image given the sentence.

\section{Experimental Setup}

In this section, we describe our setup to evaluate cross-modal influence in pretrained vision-and-language BERTs.
Our code is available online.\footnote{\url{https://github.com/e-bug/cross-modal-ablation}.}

\subsection{Evaluation data}
We use the validation split of the \entities{} dataset~\cite{flickrentitiesijcv}, a subset of Flickr30k~\cite{flickr30k}.
This dataset contains human-annotated alignments between gold image regions and phrases in image captions.

The validation set contains $1{,}000$ images, each associated to $5$ English sentences.
A sentence contains on average $2.89$ phrases.
Each phrase is aligned with $1.53$ objects on average; $78\%$ of phrases are linked to a single object.
In total, this results in $14{,}433$ phrase--object(s) data points.

\subsection{Models}
We evaluate five different architectures: LXMERT~\cite{tan2019lxmert} and ViLBERT~\cite{NEURIPS2019_c74d97b0} (dual-stream); VL-BERT~\cite{Su2020VL-BERT:}, VisualBERT~\cite{li2019visualbert} and UNITER~\cite{chen2020uniter} (single-stream).
These models extend the BERT architecture to multimodal data and tasks by supplementing the text input with image features.
Single-stream models process image and text jointly with a single Transformer encoder, while dual-stream ones first encode each modality separately and then make them interact. 
We reuse the model weights from our previous \emph{controlled} study \cite{bugliarello-etal-2021-multimodal}.\footnote{\url{https://github.com/e-bug/volta}.}

Starting from BERT, each model was pretrained on Conceptual Captions~(CC;~\citealt{sharma-etal-2018-conceptual}) for $10$ epochs.
The input images were provided as $36$ regions of interest extracted using a Faster R-CNN~\cite{NIPS2015_14bfa6bb} pretrained on Visual Genome~\cite{Anderson2017up-down}.
Each model was pretrained on three objectives: masked language modelling (MLM), masked region classification with KL divergence (MRC-KL), and image--text matching~\cite{chen2020uniter}.
We note that since each model used the same visual and textual vocabularies, as well as the same context length, cross-entropies and other entropy-based measures are comparable between model architectures.

In addition, as a control model, we adapt BERT to the same multimodal distribution by continued pretraining with MLM  on Conceptual Captions for $5$ epochs.\footnote{This results in the same number of MLM updates observed by vision-and-language models in their pretraining.}
We denote this version as BERT$_{CC}$.

\paragraph{Compute infrastructure}
Our experiments were run on a shared computing infrastructure.
Ablation setups were measured on CPU, with an average runtime of $9$ minutes.
When we pretrain \vl BERTs, we use the same hyperparameters as in our controlled study~\cite{bugliarello-etal-2021-multimodal}.
Pretraining takes $5$ days, corresponding to $10$ epochs of CC, on one NVIDIA Titan RTX GPU card.

\subsection{Prediction Tasks}
The prediction tasks mirror two of the training tasks, MLM and MRC-KL, with a few subtle but crucial differences in masking and ablation.
In the following, we denote the input textual tokens as $\mathbf{w}=\langle w_1, \dots, w_T\rangle$ and the input image regions as $\mathbf{v}=\langle \mathbf{v}_1, \dots, \mathbf{v}_K\rangle$.
In addition, we use $\mathbf{m}$ to refer to the set of masked indices, either in the textual or visual modality, and $\textbackslash\mathbf{m}$ for its complement.

\paragraph{Masked language modelling}

The MLM task is to predict the identity of a set of masked tokens $\mathbf{w}_\mathbf{m}$, given unmasked tokens $\mathbf{w}_{\textbackslash \mathbf{m}}$ and visual context $\mathbf{v}$:
\begin{equation}\label{eq:mlm}
	{\text{MLM}}(\mathbf{m}, \mathbf{w}, \mathbf{v}; \theta) = -\sum_{i\in\mathbf{m}}\log_2 \mathbb{P}_\theta(\mathbf{w}_i | \mathbf{w_{\textbackslash m}}, \mathbf{v}),
\end{equation}
where $\theta$ denotes a model's parameters.

During pretraining, tokens are masked at random, using the special \texttt{[MASK]} token.
During evaluation, only tokens corresponding to a specific grounded phrase are masked for prediction.

When ablating visual inputs (as opposed to masking), the ablated regions are replaced with the average feature vector (calculated over the evaluation dataset).
This keeps the visual inputs in-distribution, but renders them uninformative.
During \objabl, all regions that overlap with the gold region are ablated, using the intersection over target (IoT) function defined in \cref{eq:iot} below, with an overlap threshold of $\tau=0.5$.

\paragraph{Masked region classification}
The MRC task is to predict the object class of a masked visual region $\mathbf{v}_i$ given unmasked visual context $\mathbf{v_{\textbackslash m}}$ and tokens $\mathbf{w}$.
The MRC-KL variant~\cite{li2019visualbert} measures the KL-divergence of the predicted distribution rather than the cross-entropy against a single object class.
For each masked region $\mathbf{v}_i$ linked to a phrase, MRC-KL is computed as follows:
\begin{equation}\label{eq:mrckl}
	\resizebox{1\hsize}{!}{$
	{\text{MRC-KL}}(\mathbf{w}, \mathbf{v}_i; \theta) = {\text{KL}}(\mathbb{P}_g(\mathbf{v}_i)||\mathbb{P}_\theta(\mathbf{v}_i|\mathbf{w}, \mathbf{v_{\textbackslash m}})),
	$
}
\end{equation}
where $\mathbb{P}_g$ is the target object distribution and $\mathbb{P}_\theta$ is the distribution predicted by the model.
During training, the visual annotations of object classes $\mathbb{P}_g$ are given by the \faster{} object detector, which has a set of $1{,}601$ object classes.
\newcite{chen2020uniter} showed that MRC-KL leads to better downstream performance compared to MRC with cross-entropy (MRC-XE).

When a given region is masked for prediction, other regions that overlap with it are also masked out, according to the IoT function (\cref{eq:iot}). 
In practice, masking for prediction involves replacing the input region vector with a vector of zeros.
When performing textual ablation, the ablated tokens are replaced with \texttt{[MASK]} tokens.\footnote{For the \allabl setup, we also investigated using one \texttt{[MASK]} token but found no difference in model behaviour.}

During our evaluation procedure, we only evaluate region predictions for the target objects that are mentioned in the text.
As the target region, among the regions masked by the IoT function, we use the (\faster-generated) input region that has the highest overlap with the gold one, according to intersection over union (IoU; \cref{eq:iou}).

\subsection{Region Overlap Calculations}
Object detectors often output regions that overlap with each other.
These regions may leak visual information and hence need to be masked in our diagnostic procedure.
In particular, there are two points where this issue arises:
first, in language-for-vision, when deciding which regions overlap with the target region of prediction;
secondly, in vision-for-language \objabl, when deciding which regions overlap with the gold region.

\paragraph{Intersection over union}
Visual region overlap is most commonly calculated using intersection over union (IoU;~\citealt{10.1007/s11263-009-0275-4}).
Let $\mathbf{b}_i$ and $\mathbf{b}_j$ be the bounding boxes corresponding to regions $\mathbf{v}_i$ and $\mathbf{v}_j$, respectively.
IoU is computed as:
\begin{equation}\label{eq:iou}
	\text{IoU}(\mathbf{b}_i, \mathbf{b}_j) = \frac{|\mathbf{b}_i \cap \mathbf{b}_j|}{|\mathbf{b}_i \cup \mathbf{b}_j|}.
\end{equation}

Our models were pretrained by co-masking any region that overlapped with the randomly chosen one
by more than the threshold of $\tau=0.4$.

\paragraph{Intersection over target}
A shortcoming of IoU is that it may not catch large regions that largely overlap small target regions, potentially leaking information.
To avoid this, we define the more aggressive intersection over target (IoT) measure:
\begin{equation}\label{eq:iot}
	\text{IoT}(\mathbf{b}_i , \mathbf{b}_j) = \frac{|\mathbf{b}_i \cap \mathbf{b}_j|}{|\mathbf{b}_i|}.
\end{equation}
To find co-ablated regions during \objabl, we use IoT with a threshold $\tau=0.5$ (as originally recommended by \citealt{10.1007/s11263-009-0275-4} for IoU).

\section{Cross-Modal Input Ablation Results}

 \begin{figure}[t]
 	\centering
 	\includegraphics[width=\linewidth, trim={0cm 0cm 0cm 0cm}, clip]{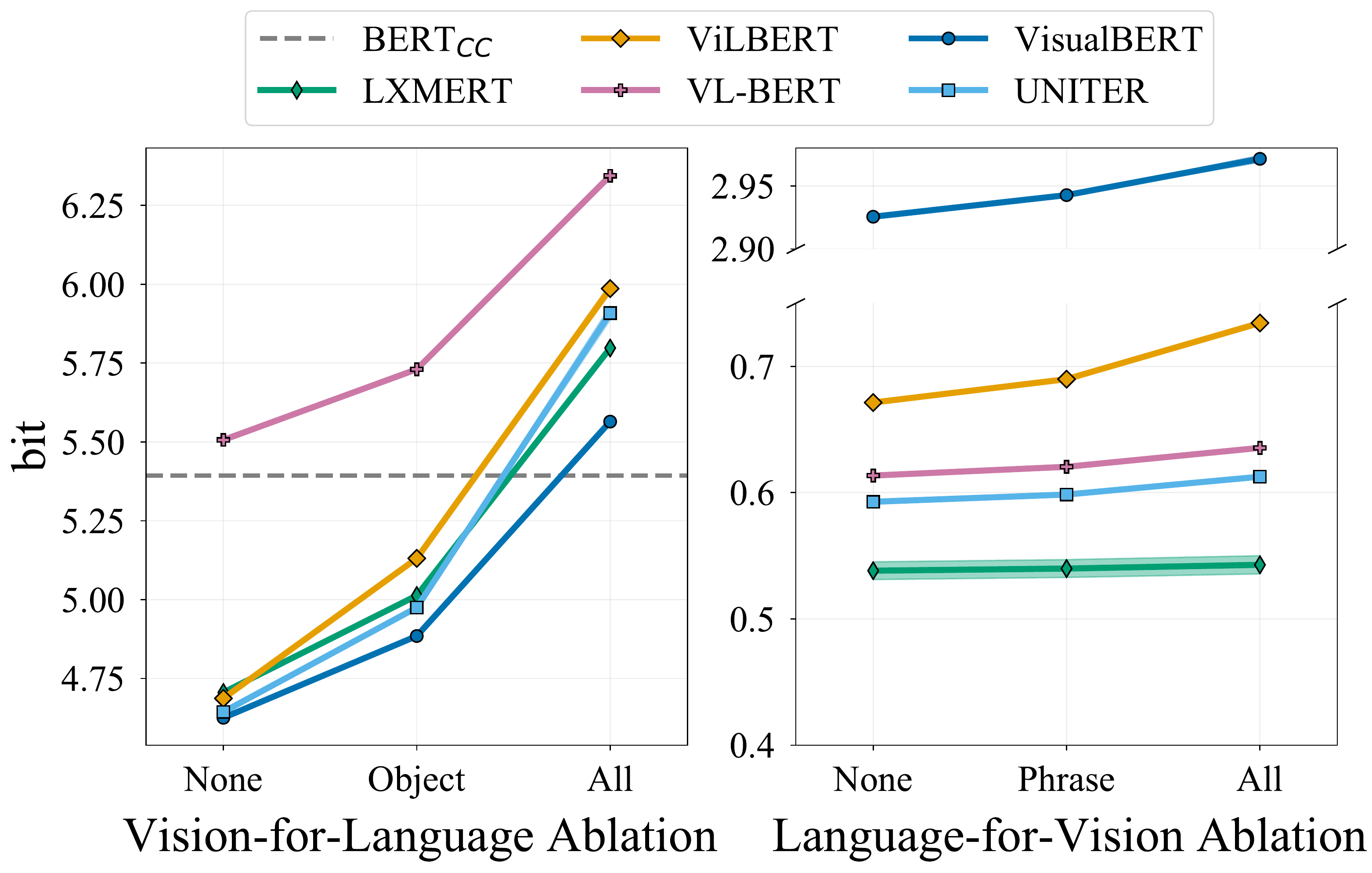}
 	\caption{Left:~Masked language modelling loss for different ablation settings: no ablation, ablation of the grounding object, ablation of the full image.
     Right:~Masked region classification with KL divergence loss for different ablation settings: no ablation, ablation of the region-aligned phrase, ablation of the full text.
     The mean performance of 10 different initial seeds is shown for each model; variance around the mean is too small to be seen for most models.
     Models suffer when visual input, but not textual input, is removed.
   }
 	\label{fig:standard_mlm_mrckl}
 \end{figure}%

\paragraph{Vision-for-Language diagnostic}

\cref{fig:standard_mlm_mrckl}~(left) shows the performance of the five models
in different types of visual input ablation.
The models perform best when
\textbf{None} of the visual input is ablated, and perform worse when the \textbf{Object} is ablated, indicating that the models are missing information about the grounding object region.
Models are most affected when removing \textbf{All} of the visual input, which is unsurprising.
However, the effect of \objabl~is relatively small, compared to the effect of \allabl, which is counter to the expectation that models should use grounding object information rather than general visual context.

\begin{figure}[t]
 	\centering
 	\includegraphics[width=1.1\linewidth, trim={0cm 0cm 0cm 0cm}, clip]{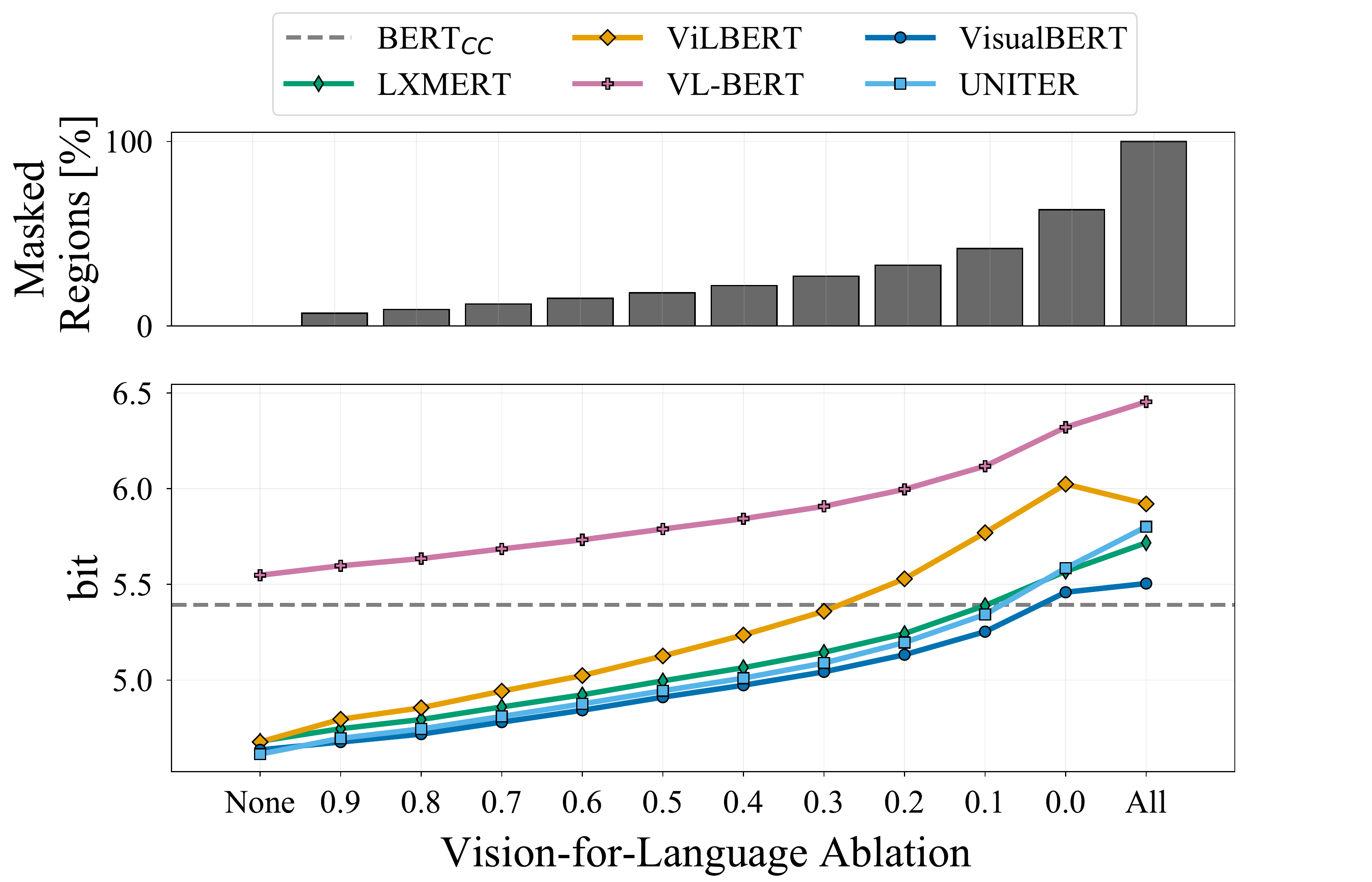}%
 	\caption{MLM performance across different visual masking thresholds, leading to increased co-masking of regions overlapping with the ablated \textbf{Object}. $\tau=0.5$ corresponds to \objabl in \cref{fig:standard_mlm_mrckl}.}
 	\label{fig:standard_mlm_thr}
\end{figure}

In \cref{fig:standard_mlm_thr}, we check whether these results are due to visual leakage of \object information into the general visual context.
We examine the effect of masking the visual \object~more or less aggressively, by changing the overlap threshold $\tau$ that triggers co-masking.
Increased masking leads to worse performance on \objabl,
and at lower thresholds for masking, the relative contribution of \textbf{Object} information is higher than that of the additional general visual context in \allabl.

Comparing models, the single-stream VisualBERT is least affected by ablating the visual input, while the dual-stream ViLBERT is most affected.
The behaviour of VL-BERT is an unexpected outlier that warrants future investigation, given that it performs similarly to other models in downstream tasks~\cite{Su2020VL-BERT:,bugliarello-etal-2021-multimodal}.

The language-only BERT model has a loss of~7.44 bits on this evaluation set, which drops to~5.39 bits after pretraining on the Conceptual Captions dataset (BERT$_{CC}$): there is a strong effect of the language domain.
Overall, when visual input is provided, the V\&L BERTs show better performance than the language-only baselines, but their performance degrades below domain-adapted unimodal performance when visual input is withheld.
{We conclude that these models have indeed learned to use vision-for-language.}

\paragraph{Language-for-Vision diagnostic}
On the right of \cref{fig:standard_mlm_mrckl} are the results of ablating language when predicting visual object categories.
Compared to the effect of ablating visual inputs for language modelling, where the performance drops by up to 25\% during \allabl, we see only minimal effects of ablating text. 
According to paired-difference t-tests, all models' ablated performances (both \textbf{Phrase} and \allabl) differ significantly from the standard model (\noneabl), with $p<0.05$.
However, the practical effects of ablation are very small.
Ablating the reference phrase leads to a increase in KL-divergence of 0.5\%--3\%, while ablating \textbf{All} text increases the KL-divergence between~1\%--10\%, relative to \noneabl.
In both cases, the least affected model is LXMERT, and the most affected model is ViLBERT.
In summary, this analysis shows that these {V\&L models use language-for-vision much less than they use vision-for-language.}

\section{Diving into Language-for-Vision}

The previous section showed that V\&L BERTs are more sensitive to ablated visual inputs than language inputs.
This behaviour could be due to several factors, including differences in model design and initialisation, pretraining objectives, and issues with the quality of the silver labels provided by \faster. Here, we analyse how these factors affect language-for-vision recruitment. 

\subsection{Model Design}
In theory, particular model architectures could favour the transfer and cross-modal use of features in one direction more than the other.
Overall, our results show all of the evaluated model designs resulting in quite similar, asymmetric, behaviour.

Interestingly, however, the two dual-stream architectures, LXMERT and ViLBERT, behave very differently.
LXMERT achieves the lowest MRC-KL value, and it is not affected by the presence of textual inputs.
ViLBERT, on the other hand, is the model that most relies on language when predicting visual features.
We believe that the main reason behind the difference in performance might be due to their different vision streams: ViLBERT directly maps visual features onto cross-modal layers, while LXMERT first processes them in a $5$-layer vision-only stream.
We hypothesise that these five layers are enough for LXMERT to learn visual representations to solve the MRC-KL task, without langauge.
Secondly, the lack of a vision-only stream in ViLBERT makes it asymmetric, which might explain its greater recruitment of language for vision.
Conversely, all the other models have approximately the same number of parameters for both modalities.

Finally, VisualBERT performs much worse than all the other models on MRC-KL.
We hypothesise this is because it is the only model that does not encode spatial information from the bounding boxes corresponding to the input visual features.

\subsection{Pretraining Loss Function}
We investigate whether pretraining with a different vision objective leads to different behaviour in language-for-vision recruitment.
In particular, predicting object class distributions, using MRC-KL, may be easier if the long-tail of object classes is relatively similar between (non-masked) regions within an image, since this would make visual contextual features highly informative.
Replacing the KL divergence objective with a cross-entropy objective forces the model to predict only the most likely object class for each region, which is expected to be more dissimilar across the image.

We pretrain a UNITER encoder using a weighted masked region classification with cross-entropy (MRC-XE) objective~\cite{tan2019lxmert}, where the weights are given by the probability assigned by the object detector to the most likely class.
We find that pretraining with this vision objective does \emph{not} lead to any change in model behaviour: Ablating the reference phrase leads to a increase in XE of 1\%, compared to the standard model (\noneabl); ablating \textbf{All} text increases it by~4\%, in line with the KL results in~\cref{fig:standard_mlm_mrckl}.

\subsection{Initialisation and Pretraining Order}

\begin{figure}[t]
 	\centering
 	\includegraphics[width=\linewidth, trim={2cm 0cm 3cm 0cm}, clip]{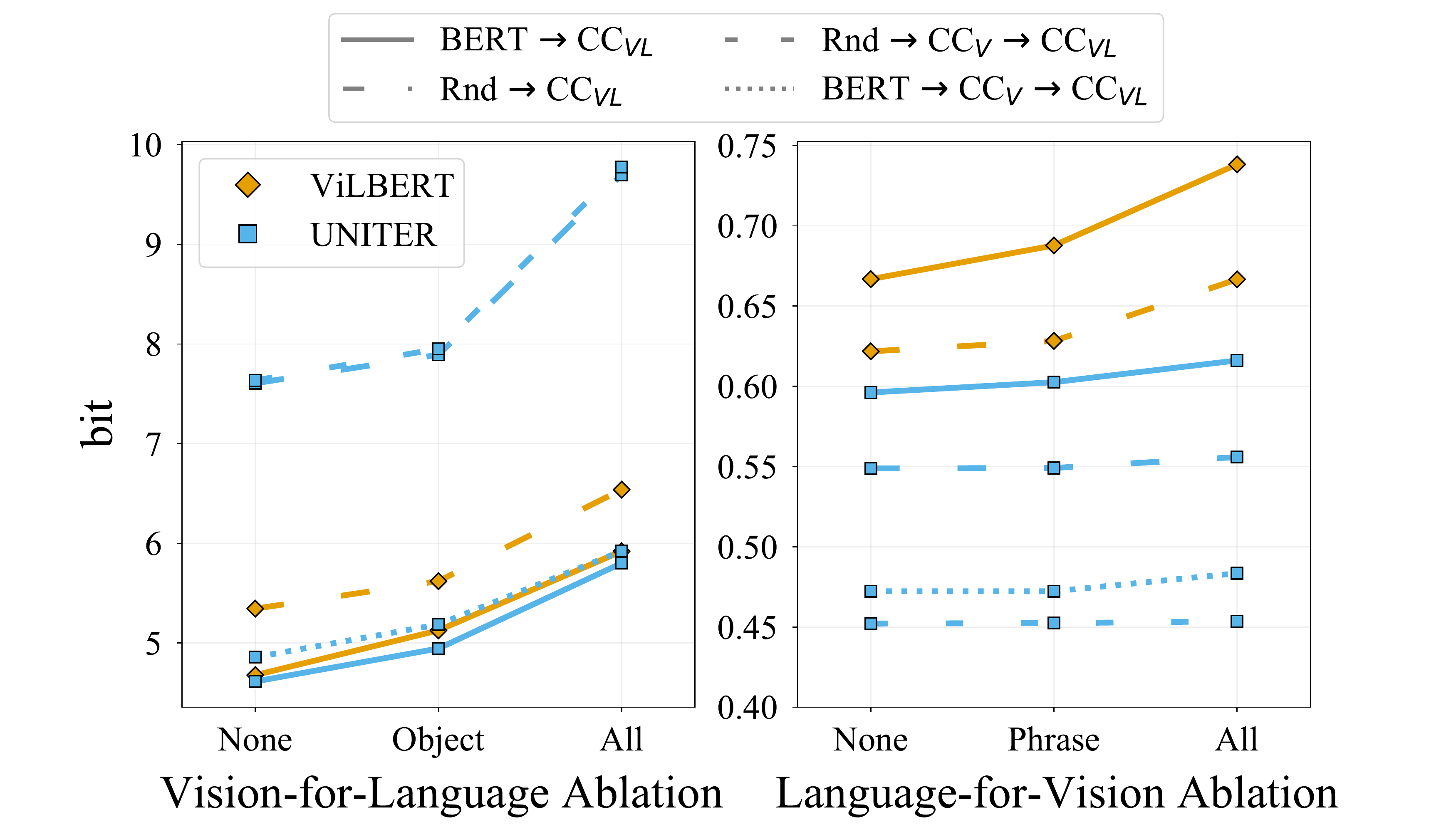}%
 	\caption{MLM and MRC-KL for different pretraining initialisations. Pretraining affects overall performance, but not ablation behaviour (lines are parallel).}
 	\label{fig:init_mlm_mrckl}
\end{figure}

All tested models share the same pretraining sequence: they are first initialised with BERT weights and then pretrained on CC to model vision and language.
We refer to this configuration as \textbf{BERT$\rightarrow$CC$_\text{VL}$}. 
Here, we ask whether the BERT initialisation leads to the asymmetric lack of language-for-vision behaviour: does the strong language modelling capability embedded in the Bert weights constrain the model from adapting towards the vision prediction task? 
In addition, we explore whether different pretraining regimes can also lead to different behaviours in \vl BERTs.
We consider the following setups:
\begin{description}[itemsep=0ex,topsep=1ex]
  \item [Rnd$\rightarrow$CC$_\text{VL}$] Instead of initialising with BERT, the models are randomly initialised and trained on CC using all three standard losses.
  \item [Rnd$\rightarrow$CC$_\text{V}$$\rightarrow$CC$_\text{VL}$] Here, randomly initialised models are first pretrained on CC using only the visual MRC-KL loss, and then pretrained on CC as usual with all three losses. This is an attempt at `vision-first' models.
  \item [BERT$\rightarrow$CC$_\text{V}$$\rightarrow$CC$_\text{VL}$] Same as above, but starting from a BERT initialisation.
\end{description}

We pretrain both ViLBERT and UNITER architectures in the \textbf{Rnd$\rightarrow$CC$_\text{VL}$} setting, and UNITER for the others.
The results for all these setups are displayed in \cref{fig:init_mlm_mrckl}, and show that (i)~pretraining from scratch increases MLM loss, compared to models initialised with BERT, in both ViLBERT and UNITER; and (ii)~pretraining with vision-first leads to lower MRC-KL loss for UNITER.
Nevertheless, our empirical findings indicate that initialisation is \emph{not} a strong factor for the observed cross-modal asymmetric behaviour.

\begin{figure}[t]
	\centering
	\includegraphics[width=\linewidth, trim={2cm 0cm 3cm 0cm}, clip]{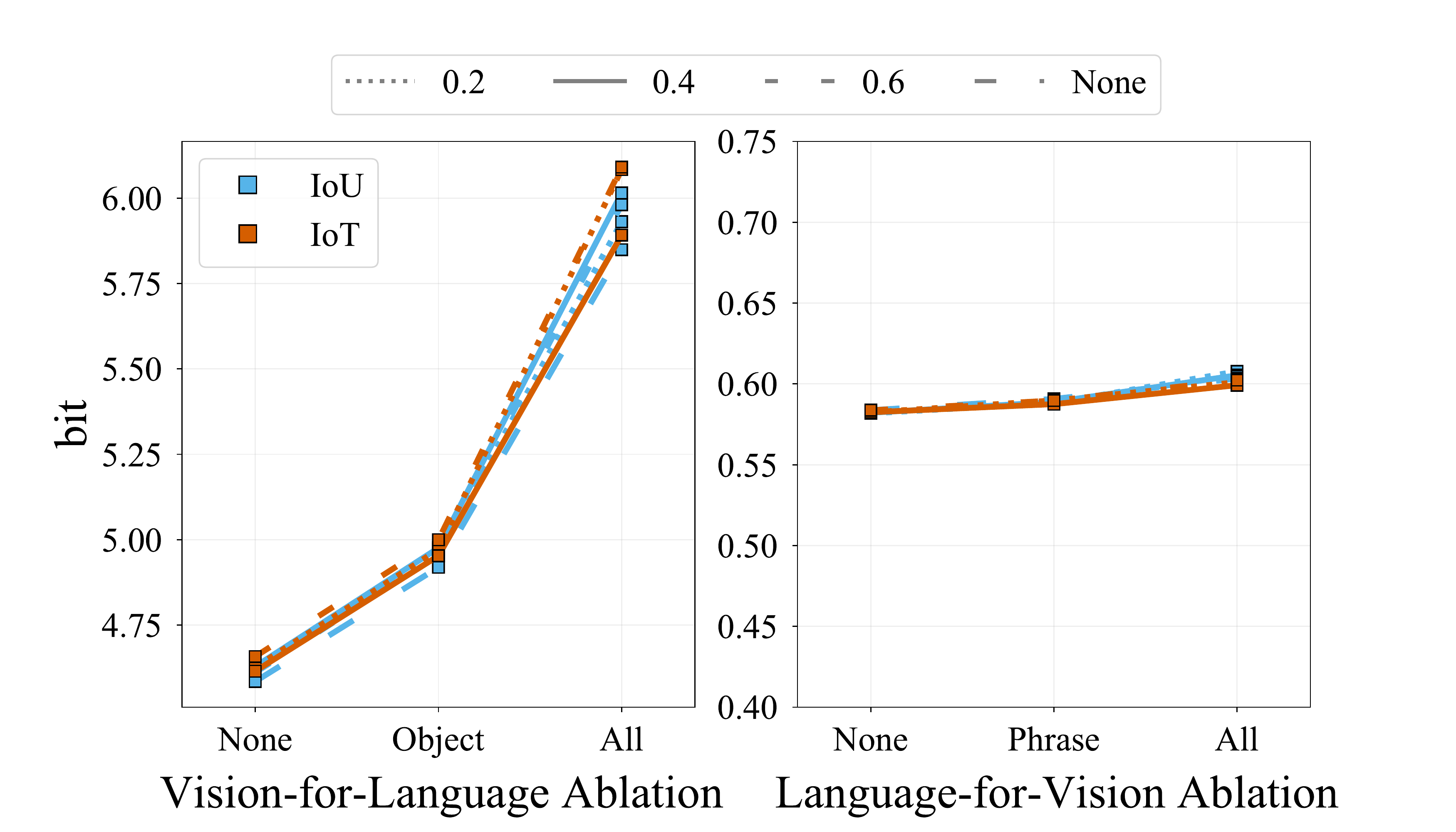}%
	\caption{MLM and MRC-KL for different pretraining thresholds for the UNITER model. Neither co-masking threshold nor co-masking function (IoU or IoT) affect model performance nor behaviour in our diagnostics.}
	\label{fig:thr_mlm_mrckl}
\end{figure}

\subsection{Leaking Visual Features}
The visual contextual features may (accidentally) provide sufficient information for correct predictions.
In particular, given that many automatically proposed regions in an image overlap, if these are not correctly co-masked with the \object region, they could leak information about the \object.

We perform an experiment where we pretrain UNITER with different overlapping thresholds $\tau\in\{0.2, 0.4, 0.6\}$ as well as without any co-masking (None), testing both IoU and IoT masking functions.
\cref{fig:thr_mlm_mrckl} shows that varying the amount of co-masking, or the masking function, does \emph{not} affect the recruitment of language-for-vision.
Our result counters the hypothesis of \citet{Lu_2020_CVPR} that co-masking \emph{``forces the model to rely more heavily on language to predict image content.''}

\subsection{Silver Object Annotations}
Finally, we examine the data used for representing the visual modality.
The models are trained and evaluated on object predictions, `silver distributions,' from \faster, which contain noise because they are automatically predicted categories. 
During evaluation, noise in the target distributions may lead to discounting of the role of language if the evaluation set contains many examples in which the target object class predictions conflict with the aligned textual features.

To examine this, we construct a subset of the \entities{} validation set, called \entsvg{}, in which the nouns of an aligned phrase -- extracted using Stanza~\cite{qi2020stanza} -- perfectly match with one of the \faster's object labels.
(Note that this does not mean that \faster{} predicted that category for that example, only that it is possible for the model to have done so.)
This results in $10{,}159$ data points out of the original $14$k.
We consider these labels, extracted from \entities{}, as human-generated ``gold'' object labels.

\paragraph{The silver distributions are noisy}
\begin{figure}[t]
	\centering
	\includegraphics[width=.9\linewidth, trim={0cm 0cm 0cm 0cm}, clip]{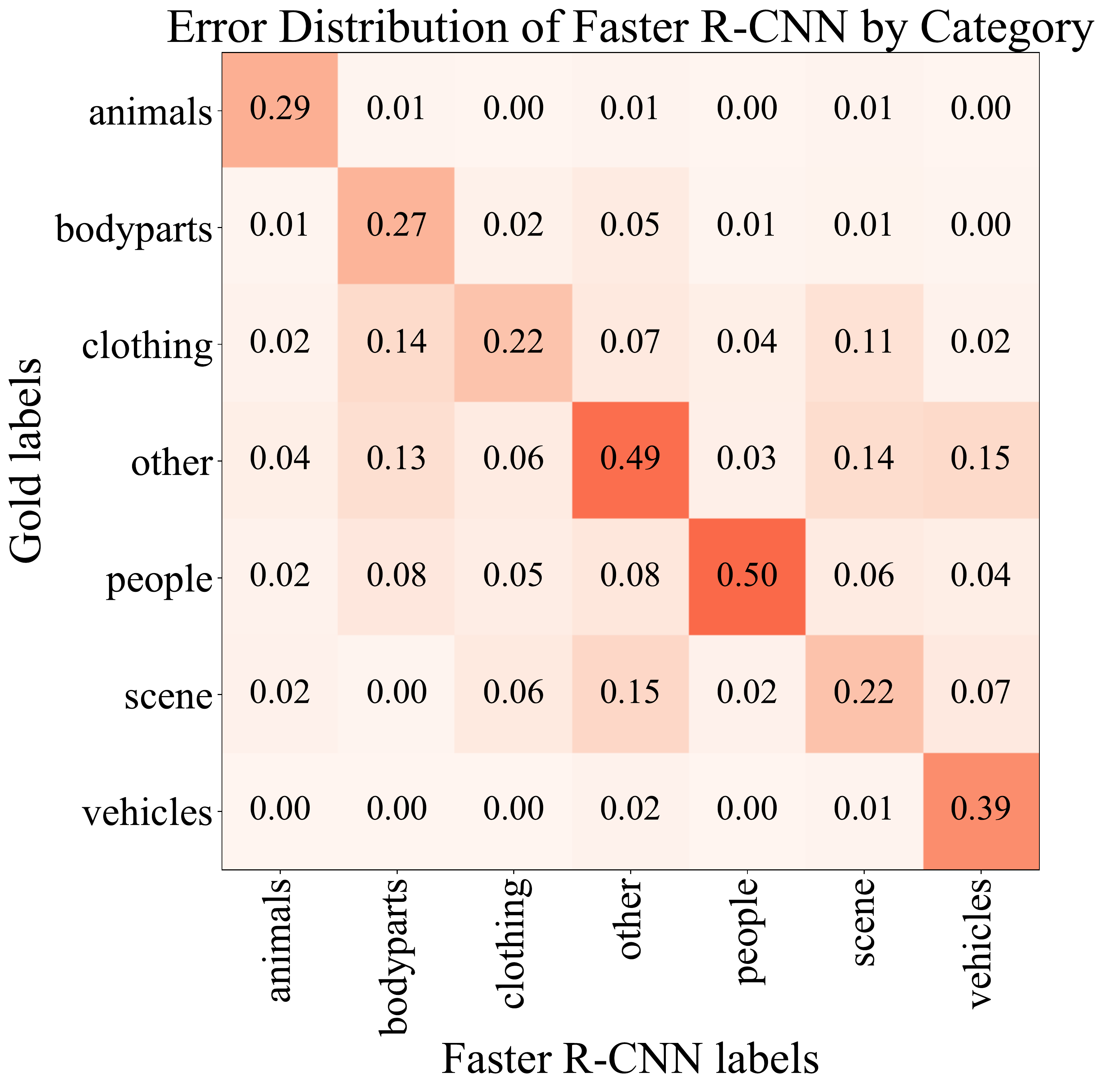}%
	\caption{Confusion matrix showing the proportion of incorrect class predictions (out of all predictions), grouped into categories. $60\%$ of \faster{} ``people'' category label predictions are incorrect (sum of column); $50\%$ of all predictions are incorrect within the set of ``people'' classes.}
	\label{fig:confusion}
	\vspace{-0.2cm}
\end{figure}

On the \entsvg{} subset, the highest-probability class from \faster{} agrees with the gold label only on $38\%$ of examples.
The gold label is in the top\nobreakdash-3 $55\%$ of the time, in top\nobreakdash-5 $64\%$, and in top\nobreakdash-10 $75\%$ of the time. 
\cref{fig:confusion} shows the distribution of errors, grouped into the higher-level categories defined in the Flickr30k Entities dataset.
\faster{} is mostly making mistakes within categories, especially within the ``people'' category, where more than half of all predictions are wrong according to the gold label.
There is also a long tail of wrong predictions not in the Flickr30k Entities object classes within the ``other'' predictions.

\begin{figure}[t]
	\centering
	\includegraphics[width=\linewidth, trim={0cm 0cm 0cm 0cm}, clip]{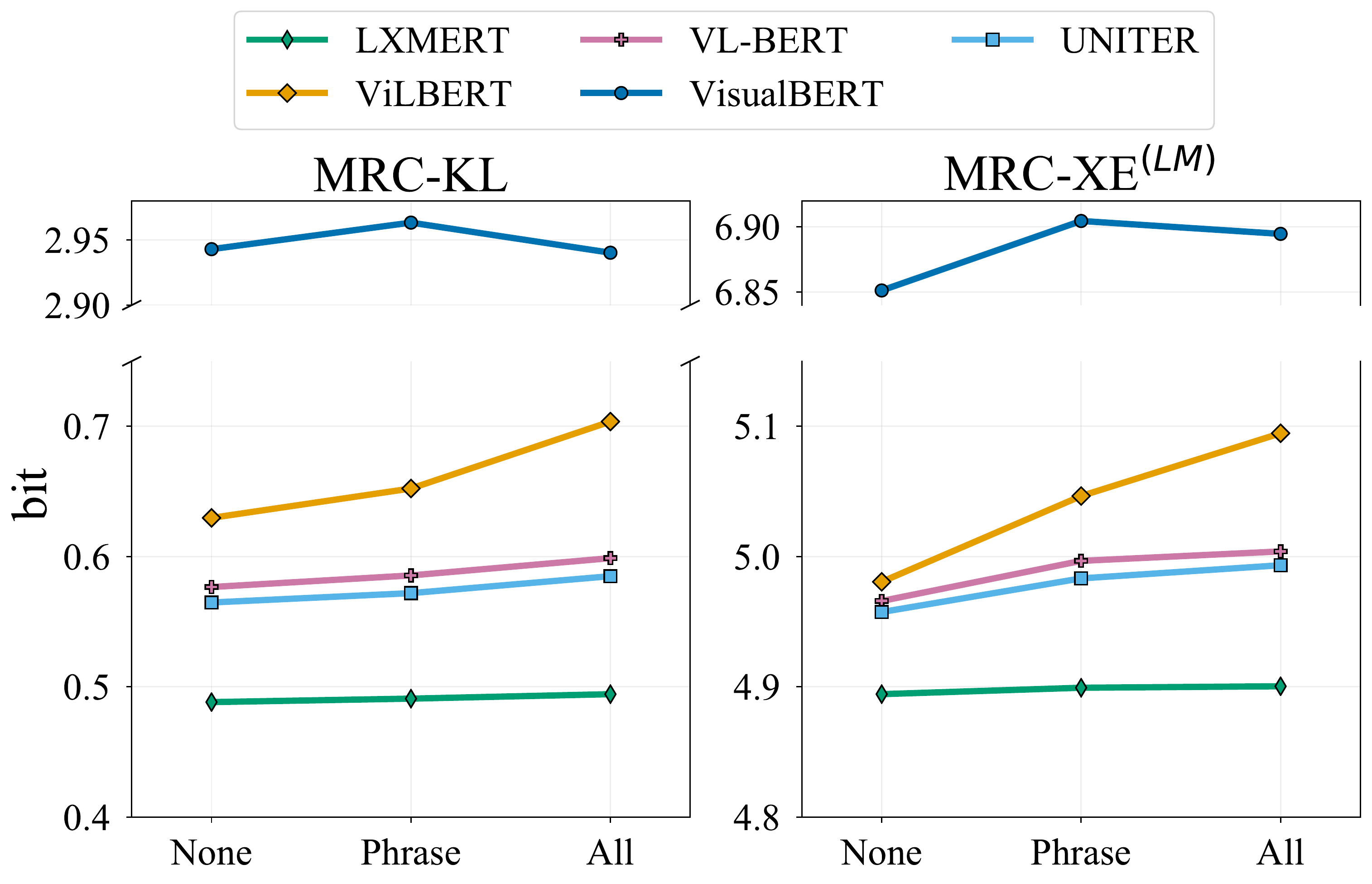}%
	\caption{\entsvg{} subset results: MRC-KL against silver distributions and MRC-XE against gold labels.}
	\label{fig:vg_mrckl_mrcxevg}
	\vspace{-0.2cm}
\end{figure}
\begin{figure}[t]
	\centering
	\includegraphics[width=\linewidth, trim={0cm 0cm 0cm 0cm}, clip]{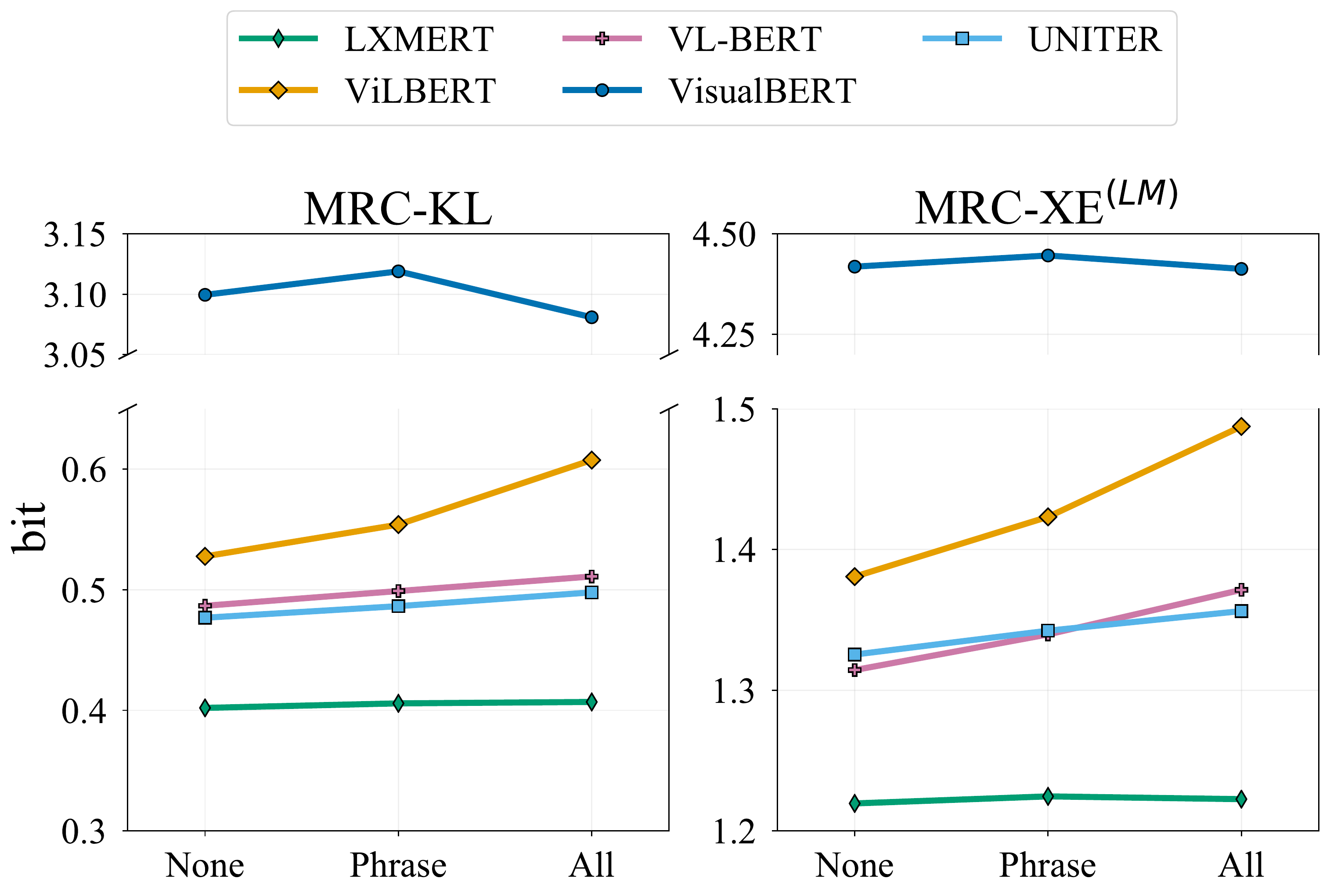}%
	\caption{MRC-KL and MRC-XE in the subset of LabelMatch where \faster's prediction match with gold labels. We observe similar patterns as in \entsvg{} showing that models do not recruit text for vision even when the predicted object classes are correct.}
	\label{fig:standard_vgpp_mlm_mrckl}
	\vspace{-0.1cm}
\end{figure}

\paragraph{Evaluating on gold labels}
Using the \entsvg{} as gold labels for evaluation does not lead to any clear difference in ablated MRC performance, as seen in \cref{fig:vg_mrckl_mrcxevg} (right). (\cref{fig:vg_mrckl_mrcxevg} (left) confirms that this subset does not behave differently from the full validation set on MRC-KL against the silver distribution labels.)
Models still do \emph{not} rely on text when predicting the (gold) region class, even though --- in the un-ablated setting --- the region label is given in the text input in this dataset.

Finally, we also measure MRC-KL on the subset of \entsvg{} in which the silver distribution mode and the gold labels agree.
This tests whether models use language for vision when the most likely class of silver distribution is correct: perhaps these instances are recognisably cleaner to the model.
However, we again find no difference in ablated performance (see \cref{fig:standard_vgpp_mlm_mrckl}).

In conclusion, even when evaluated on gold labels, we still see most models making next to \emph{no} use of textual information for visual predictions. 
This behaviour is consistent with models that have been pretrained against noisy silver data, where language inputs are not useful for prediction.\footnote{We believe it also explains why \newcite{chen2020uniter} found an advantage of pretraining with a KL-divergence loss instead of a maximum-probability cross-entropy loss: predictions made with KL divergence are more robust to noise.}

\section{Conclusion}

The \emph{cross-modal input ablation diagnostic} introduced in this paper demonstrated an asymmetry in
pretrained vision and language models: the prediction of masked text is
strongly affected by ablated visual inputs, compared to (almost) no effect of
ablating textual inputs when predicting masked image regions.
These results offer a useful check of actual model behaviour, and run counter to hypotheses assuming more balanced cross-modal activations.

We conducted several follow-up studies to better understand, and possibly ameliorate, this behaviour.
We explored (i)~pretraining the model on vision before cross-modal training; (ii)~visual leakage from overlapping image regions; and (iii)~problems with the silver target distributions produced by \faster{}.
There were no discernible effects from (i) or (ii).
With regards to (iii),
we found the silver labels to be much less reliable than expected.
We suspect that these silver target distributions are not appropriate for activating language-for-vision,
and we recommend that future V\&L models avoid them, if they aim to not only support vision-for-language.

This paper has focused on an intrinsic evaluation of pretrained models, rather than their downstream task performance.
We note that the models studied in this paper perform similarly on a number of downstream tasks~\cite{bugliarello-etal-2021-multimodal},
in line with their similar ablation behaviour.
Whether cross-modal ablation behaviour can predict downstream performance is a question left for future models with more divergent behaviour.

More broadly, the fact that the tested models showed an effect of vision-for-language but not language-for-vision may be an accumulation of model engineering for multimodal tasks that are more strongly vision-for-language, such as visual question answering or grounded reasoning.
Some recent models~\cite{li2021align,shen2021clip,hendricks-nematzadeh-2021-probing} have removed the visual component from the training objective entirely, resulting in vision-for-language architectures.
In the future, we advocate for increased work on more language-for-vision tasks,
such as text-modulated object detection~\cite{kamath2021mdetr},
in order to push pretrained vision-\emph{and}-language models to be truly, bidirectionally, cross-modal.

\section*{Acknowledgements}
{\scriptsize\euflag} We are grateful to the anonymous reviewers and members of the CoAStaL NLP group for their constructive feedback. This project has received funding from the European Union's Horizon 2020 research and innovation programme under the Marie Sk\l{}odowska-Curie grant agreement No 801199.

\bibliography{custom}
\bibliographystyle{acl_natbib}

\end{document}